\pdfoutput=1
\documentclass{article}
\usepackage[final,nonatbib]{neurips_2020}
\usepackage[utf8]{inputenc} 
\usepackage[T1]{fontenc}    
\usepackage{url}            
\usepackage{booktabs}       
\usepackage{amsfonts}       
\usepackage{nicefrac}       
\usepackage{microtype}      
\usepackage{lipsum}
\usepackage{graphicx}
\graphicspath{ {./images/} }
\usepackage{multirow}
\usepackage{floatrow}
\usepackage{caption}
\usepackage{wrapfig}
\usepackage[bookmarks=false]{hyperref}
\hypersetup{colorlinks,allcolors=black}

\newfloatcommand{capbtabbox}{table}[][\FBwidth]

\title{\textit{A Higher Purpose: Measuring Electricity Access Using High-Resolution Daytime Satellite Imagery}}

\author{
 Zeal Shah \thanks{Corresponding author.}\\
  STIMA Lab\\
  UMass Amherst\\
  \texttt{zshah@umass.edu} \\
   \And
Simone Fobi \\
  Columbia University\\
  \texttt{sf2786@columbia.edu} \\
  \And
Gabriel Cadamuro \\
  Atlas AI\\
  \texttt{gabe@atlasai.us} \\
  \And
Jay Taneja \\
  STIMA Lab\\
  UMass Amherst\\
  \texttt{jtaneja@umass.edu} \\
}

\begin{document}
\maketitle
\begin{abstract}
Governments and international organizations the world over are investing towards the goal of achieving universal energy access for improving socio-economic development. However, in developing settings, monitoring electrification efforts is typically inaccurate, infrequent, and expensive. In this work, we develop and present techniques for high-resolution monitoring of electrification progress at scale. Specifically, our 3 unique contributions are: (i) identifying areas with(out) electricity access, (ii) quantifying the extent of electrification in electrified areas (percentage/number of electrified structures), and (iii) differentiating between customer types in electrified regions (estimating the percentage/number of residential/non-residential electrified structures). We combine high-resolution 50 cm daytime satellite images with Convolutional Neural Networks (CNNs) to train a series of classification and regression models. We evaluate our models using unique ground truth datasets on building locations, building types (residential/non-residential), and building electrification status. Our classification models show a 92\% accuracy in identifying electrified regions, 85\% accuracy in estimating percent of (low/high) electrified buildings within the region, and 69\% accuracy in differentiating between (low/high) percentage of electrified residential buildings. Our regressions show $R^2$ scores of 78\% and 80\% in estimating the number of electrified buildings and number of residential electrified building in images respectively. We also demonstrate the generalizability of our models in never-before-seen regions to assess their potential for consistent and high-resolution measurements of electrification in emerging economies, and conclude by highlighting opportunities for improvement.
\end{abstract}


\section{Introduction}
\label{sec:intro}

The United Nations Sustainable Development Goal 7 (SDG 7) is aimed at achieving universal access to affordable, sustainable, and reliable energy by 2030. Accurate assessment and tracking of electrification and its extent is crucial for prioritizing investments in infrastructure development.
Although monitoring electrification is a key component for integrated electricity systems planning, geospatial data on electrification is often scarce and inaccurate or even unavailable. Particularly in resource-constrained settings, traditional methods of collecting data on electrification require on-the-ground surveys that are expensive and thus infrequent. A common alternative approach to measurement is via satellite-recorded data of nighttime lights (NL) illumination \cite{hrea,gdessa,gridfinder}. However, these NL data are noisy due to their high susceptibility to external factors like lunar illuminance, cloud cover, stray lights, and the scan angle of the satellite \cite{eog_reliability}. Additionally, performance of detection suffers in rural areas, where typically dim lights may get obscured by external noise \cite{eog_reliability,littlelight}. Furthermore, NL data have a relatively low spatial resolution of 15 arc-seconds ($\approx$ 450 meters at the equator), providing limited detail.

In this work, we present an approach to monitor electrification using high-resolution daytime satellite imagery. Our work harnesses two innovation trends: first, advancements in the field of remote sensing and imaging technology has enabled collection of high-resolution (30-50cm) daytime satellite imagery across the globe. Second, innovation in the field of computer vision has led to diverse neural network techniques for extracting meaningful information from imagery.

We assemble a unique ground truth dataset that provides geo-coordinates, their electrification status, and building type of all structures in the country of Kenya. We employ this unique ground truth dataset to train CNN models using daytime satellite images to: (i) measure electricity access, (ii) quantify the number of electrified structures, and (iii) estimate the number of electrified structures that are (non-)residential. Furthermore, in this work, we focus on the generalizability of our models for each prediction objective by evaluating performance on images from never-before-seen Kenyan counties (herein, the out-of-sample test set). Our prediction techniques can serve as a tool to assist energy system planners, utilities, and other stakeholders in efficiently identifying areas for expanding electrification, choosing the right energy supply systems for new electrification, and appropriately sizing these systems anticipating future demand growth. Our methods address these challenges by leveraging the ever-growing corpus of high-resolution daytime satellite imagery. 

\section{Related Work}
\label{sec:related}
\textbf{Measuring electricity access and quantifying extent of electrification}: Traditional methods for collecting electrification data -- surveys -- are expensive, and the publicly available survey data from international organizations like the World Bank \cite{wb_elecaccess} have low-spatial resolution making it impossible to obtain granular insights into progress of electrification. Moreover, technological resource constraints have shown to have limited crowd-sourcing data collection efforts in developing countries \cite{crowdsource_grid}. Such limitations of bottom-up approaches have led researchers to collect data in a top-down fashion using remote sensing. Remotely sensed nighttime lights (NL) data has been the most preferred dataset for electrification studies and has been used in identifying electricity access \cite{hrea}, mapping high and medium voltage power lines \cite{gridfinder}, analyzing grid reliability \cite{eog_reliability}, and estimating population without electricity access \cite{gdessa}. Our work focused on identifying regions with(out) access to electricity is closest to \cite{hrea}, and our work on quantifying the number of electrified structures shares similar goals with \cite{gdessa}, though both the studies rely solely on the NL dataset, which is low resolution and highly susceptible to external noise. In this paper, we instead leverage high-resolution daytime satellite imagery to study electrification and quantify its extent, and show that our model performs significantly better than the NL-based baseline in identifying regions with access to electricity. Our work derives motivation from multiple projects that have applied CNNs to daytime satellite images for energy systems planning and measurements like microgrid siting \cite{india_microgrids}, solar PV identification \cite{solarPV}, solar production estimation \cite{deeproof}, and energy consumption prediction \cite{daytime_energy_consumption}. 

\textbf{Identifying building type}: Urban structural unit classification demonstrated in \cite{usu,usu_nigeria} closely resembles our work on predicting the density of (non-)residential type structures in daytime satellite images. However, in addition to density, we demonstrate prediction of actual number of electrified (non-)residential structures at $250m \times 250m$ resolution, which to the best of our knowledge has not been achieved using daytime satellite imagery. Furthermore, we believe that our work can potentially supplement electrification and structure type information to the publicly-available and widely used building footprints and population datasets \cite{googlebuild,msftbuild,hrsl}.

\section{Methodology}
\label{sec:methodology}

\subsection{Datasets and data pre-processing}
\label{sec:data}

\begin{figure*}[t]
\begin{tabular}{|c|c|c|c|}
No building & Unelectrified & Electrified & Electrified \\
& & residential & non-residential \\
\includegraphics[width=0.2\textwidth, bb=0 0 500 500]{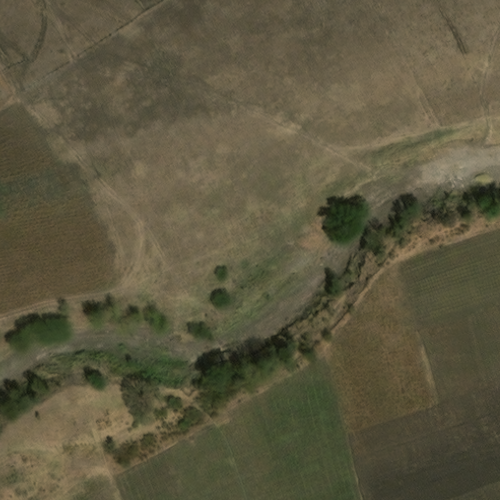} &  
\includegraphics[width=0.2\textwidth, bb=0 0 500 500]{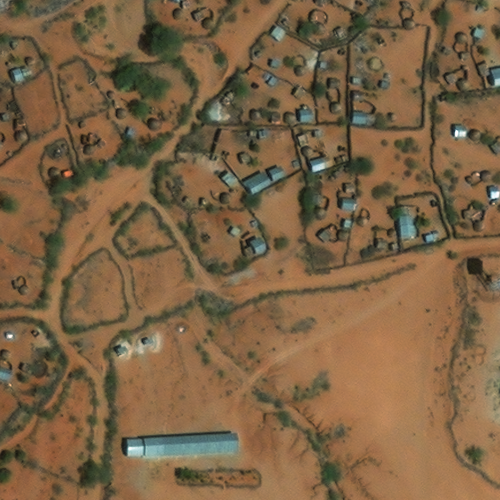} & 
\includegraphics[width=0.2\textwidth, bb=0 0 500 500]{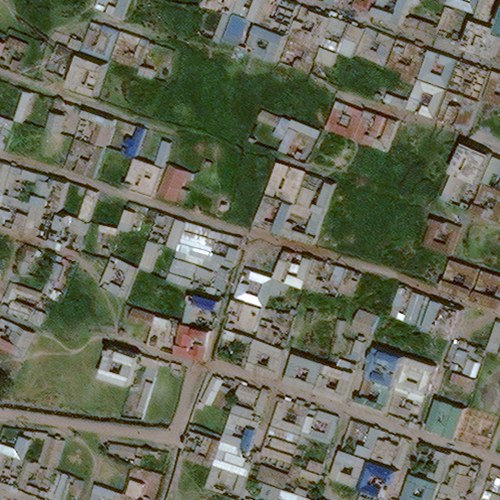} & 
\includegraphics[width=0.2\textwidth, bb=0 0 500 500]{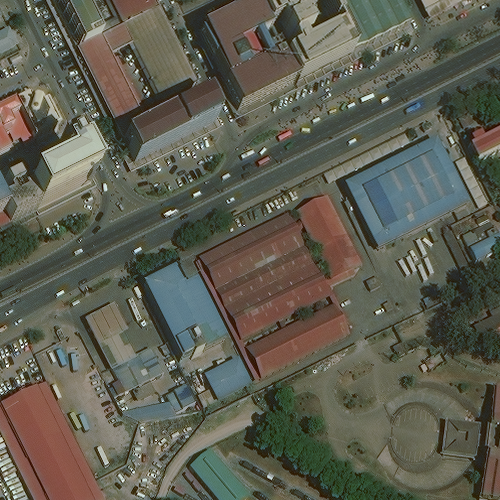} 
\\
\includegraphics[width=0.2\textwidth, bb=0 0 500 500]{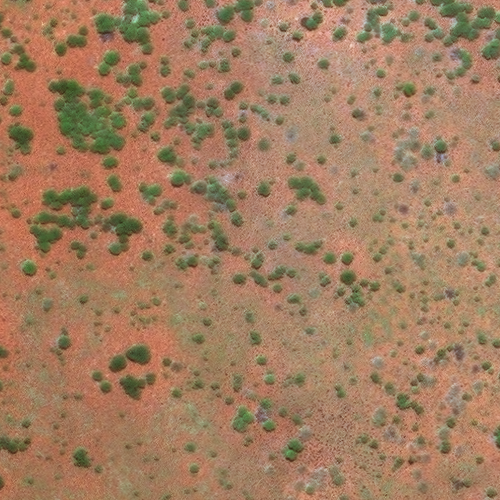} & 
\includegraphics[width=0.2\textwidth, bb=0 0 500 500]{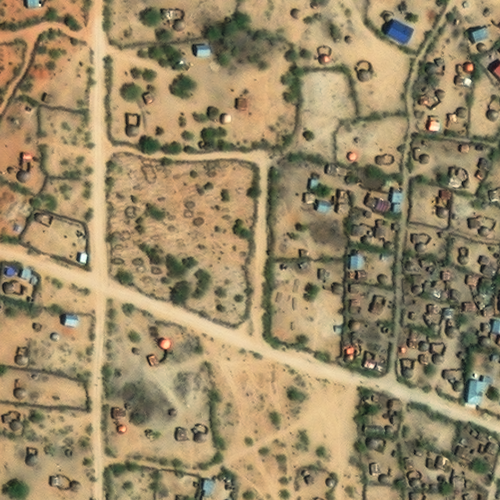} & 
\includegraphics[width=0.2\textwidth, bb=0 0 500 500]{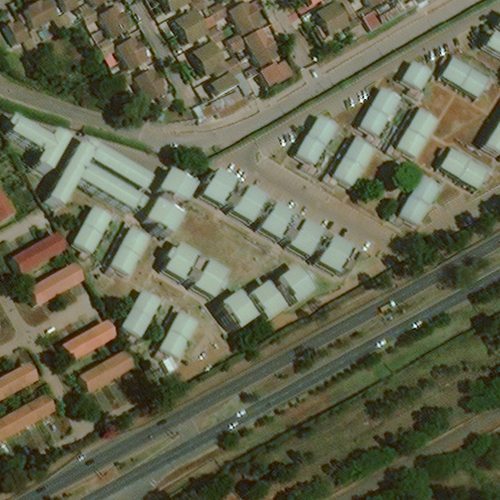} & 
\includegraphics[width=0.2\textwidth, bb=0 0 500 500]{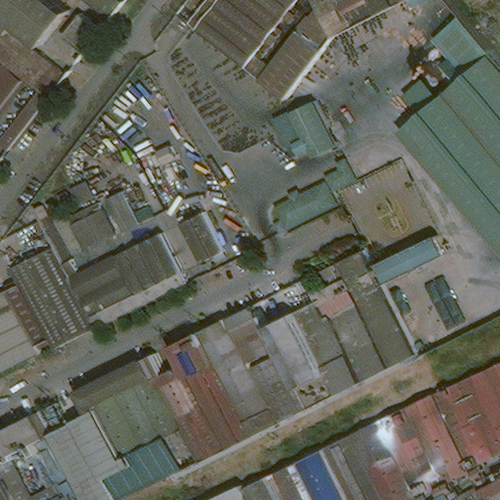}  \\
\end{tabular}
\caption{Sample image tiles from the dataset showing different electrification status}
\label{images_map}
\end{figure*}

\begin{wrapfigure}{R}{0.5\textwidth}
  \begin{center}
    \includegraphics[width=0.65\textwidth, bb=0 0 860 870]{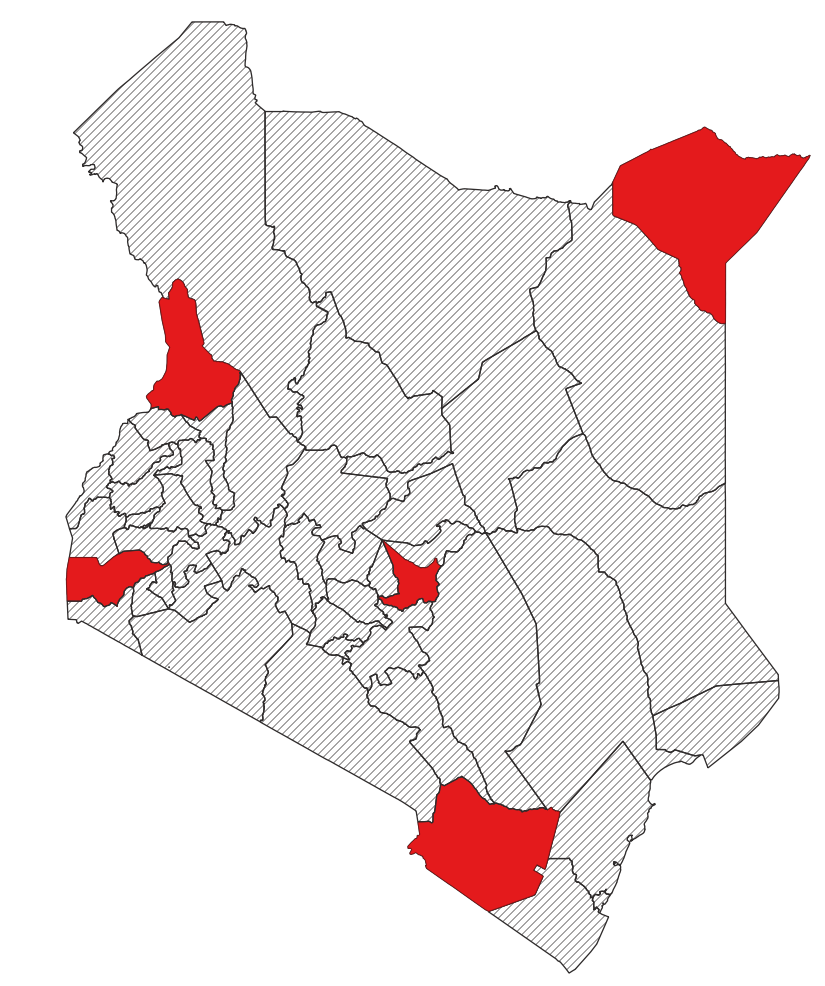}
  \end{center}
  \caption{Geographical splitting of Kenyan counties. Dark red regions represent 5 out-of-sample counties. Hashed regions represent 42 counties used for training, validation, and in-sample testing of prediction models.}
\end{wrapfigure}


We use 50cm resolution daytime images of Kenya from DigitalGlobe (DG) \cite{DG}. For ground truth, we use two structure location datasets in Kenya -- the first contains locations and types of electrified utility customers and the second contains all building locations as recorded for the Kenya National Electrification Strategy (KNES) \cite{KNES}. 

\textbf{Extracting daytime image tiles:} DG's snapshot of Kenya consists of 7000 images of size $10km \times 10km$ at $50cm$ resolution. For ease of training and evaluation, we divided Kenya into a grid of tile size $500px \times 500px$ i.e. $250m \times 250m$, and extracted the corresponding image tiles from the large-sized DG images (as seen in Figure \ref{images_map}(a)). Different DG images were captured on different dates and so we extracted year of capture for each image tile which ranged from 2012 to 2017. In the end, we were left with approximately 9 million image tiles covering Kenya.

\textbf{Processing labels:} The utility dataset provides locations and types of electrified customers: residential, commercial, and industrial. For simplicity, all the commercial and industrial customers were labeled as non-residential customers. We grouped all customers sharing the same location coordinates and same connection type into one electrified structure, and then computed the total number of electrified structures and total number of electrified (non-)residential structures in every image tile. The KNES dataset provided the total number of structures inside each image tile. Finally, for a given image tile, we obtain the total number of unelectrified structures by simply subtracting the total number of electrified structures from the total number of structures.

\textbf{Dealing with temporal mismatch:} The utility and KNES surveys were conducted in 2015-16 while capture dates for DG's daytime images ranged from 2012 to 2017, introducing a temporal mismatch. To minimize the impact of this mismatch and avoid misleading results, we only used images captured between 2014 and 2017. This reduced the total number of image tiles to 6.3 million with 5.4 million tiles containing no structures, 0.2 million containing at least one electrified structure, and 0.7 million containing only unelectrified structures.   
\subsection{Prediction objectives}
\label{sec:objectives}

We present the model training approaches we employed for the objectives discussed in Section \ref{sec:intro}. 

\textbf{Objective A: Measuring electricity access.} To measure electricity access, we perform a 3-class classification task, identifying areas containing only unelectrified structures, at least one electrified structure, or no structures at all. Such spatially distributed information about electricity access can help energy system planners and utilities identify the next set of areas to electrify. 

\textbf{Objective B: Quantifying the extent of electrification}. For electrified areas, we measure i.) percentage of electrified structures within an image using a binary classification approach, and ii.) the number of electrified structures within an image using a regression. The former provides planners with a relative measure of how many electrified structures the image contains thereby identifying areas with lower rates of electrification, while the latter provides an absolute measure (noting that total building numbers in an image are not available at inference). Both offer different yet complementary pieces of information. For the binary classification task on percent of electrified structures, a threshold is used, denoting images with low ($<=25\%$) or high ($>25\%$) percentage of electrified structures.

\textbf{Objective C: Quantifying types of electrified structures.} Residential customers have different energy demand profiles from non-residential customers like, industries. Therefore, information about the percentage and number of different types of electrified customers in an image can help planners estimate energy demand of those regions. Following the same approach as for Objective B, we predict i.) percent of electrified residential structures using a binary classification applying a threshold with low and high percentages being $<=25\%$ and $>25\%$ respectively, and ii.) using a regression model we estimate the actual number of electrified residential structures in images.

\subsection{Model training}
\label{sec:training}

We split our entire dataset into train, validation, in-sample test, and out-of-sample test sets. To obtain the out-of-sample test set, we divided Kenya's map into four quadrants and selected images of a single county from each quadrant plus an additional county from the center (see Figure \ref{images_map}(b)). In addition to being geographically distinct, counties in the  out-of-sample test set were selected to ensure the model gets evaluated on images with varied density of structures per image. Since the models never get to see images from counties in the out-of-sample test set, it presents a strong and challenging case for evaluating model generalizability. Data from the remaining counties was randomly assigned to the train, validation, and in-sample test sets with 70\%:20\%:10\% proportions ensuring similar distribution of structures per image per county in each set. 

A transfer learning approach was used, leveraging the VGG11 network \cite{VGG} initialized with ImageNet weights. An additional convolutional layer was added before the first VGG11 layer to accommodate input image tiles of size $500\times500$. All VGG11 layers were tuned by optimizing learning rates ($1e^{-8}$ - $1e^{-4}$), batch size (16-32 images), and training time (50-100 epochs) across all models.

\section{Results and Discussion}
\label{sec:results}

\begin{table}[t]
\begin{tabular}{ cc }  
\begin{tabular}{|c|c|c|} \hline
    \textbf{Model} & \textbf{Unelec.} &\textbf{Elec.} \\
     & \textbf{areas} & \textbf{areas} \\
    \hline
    \hline
    Baseline (HREA \cite{hrea}) & 0.98 & 0.64 \\
    \hline
    Elec. access (ours) & 0.98 & 0.75 \\
    \hline
\end{tabular} &  
\begin{tabular}{ |c|c|c| } 
    \hline
    \textbf{Label} & \textbf{In-sample} & \textbf{Out-of} \\
    &  & \textbf{sample} \\
    \hline
    \hline
    No building & 0.95 & 0.96 \\
    \hline 
    Elec. building & 0.81 & 0.73 \\
    \hline
    Unelec. building & 0.75 & 0.80 \\
    \hline
    \hline
    \textbf{Overall acc.} & 0.94 & 0.92 \\
    \hline
\end{tabular} \\
(a) & (b)
\end{tabular}
\caption{Objective A results -- (a) performance of our electricity access model relative to baseline, and (b) performance of our model in predicting images with no, electrified, and unelectrified buildings.}
\label{tab:objective1_results}
\end{table}

\begin{figure}[t]
\begin{floatrow}
\capbtabbox{%
\begin{tabular}{|c|c|c|c|}
    \hline
     & \textbf{Label} & \textbf{In-} & \textbf{Out-of-} \\
     & & \textbf{sample} & \textbf{sample} \\
    \hline
    \hline
    \multirow{4}{*}{\textbf{\shortstack[l]{Classif.\\ acc.}}} & low perc. & 0.86 & 0.92 \\\cline{2-4}
    & high perc. & 0.70 & 0.60 \\\cline{2-4}
    & \textbf{Overall} & 0.85 & 0.89 \\
    \hline
    \hline
    \textbf{Reg. $R^2$} & \textbf{} & 0.78 & 0.66\\
    \hline
\end{tabular}
}{
\caption{Objective B results -- quantifying percent/number of electrified structures. Classification accuracy and regression $R^2$ scores for in-sample and out-of-sample test sets.}
\label{tab:objective2_results}
}
\capbtabbox{%
    \begin{tabular}{|c|c|c|c|}
        \hline
         & \textbf{Label} & \textbf{In-} & \textbf{Out-of-} \\
         & & \textbf{sample} & \textbf{sample} \\
        \hline
        \hline
        \multirow{4}{*}{\textbf{\shortstack[l]{Classif.\\ acc.}}} & low perc. & 0.69 & 0.70 \\\cline{2-4}
        & high perc. & 0.65 & 0.64 \\\cline{2-4}
        & \textbf{Overall} & 0.69 & 0.68 \\
        \hline
        \hline
        \textbf{Reg. $R^2$} & \textbf{} & 0.80 & 0.55\\
        \hline
    \end{tabular}
}{
\caption{Objective C results -- quantifying percent/number of electrified residential structures. Classification accuracy and regression $R^2$ scores for in-sample and out-of-sample test sets.}
\label{tab:objective3_results}
}
\end{floatrow}
\end{figure}

\begin{figure}[t]
\centering
\begin{tabular}{cc}
\includegraphics[width=0.48\linewidth, bb=0 0 440 440]{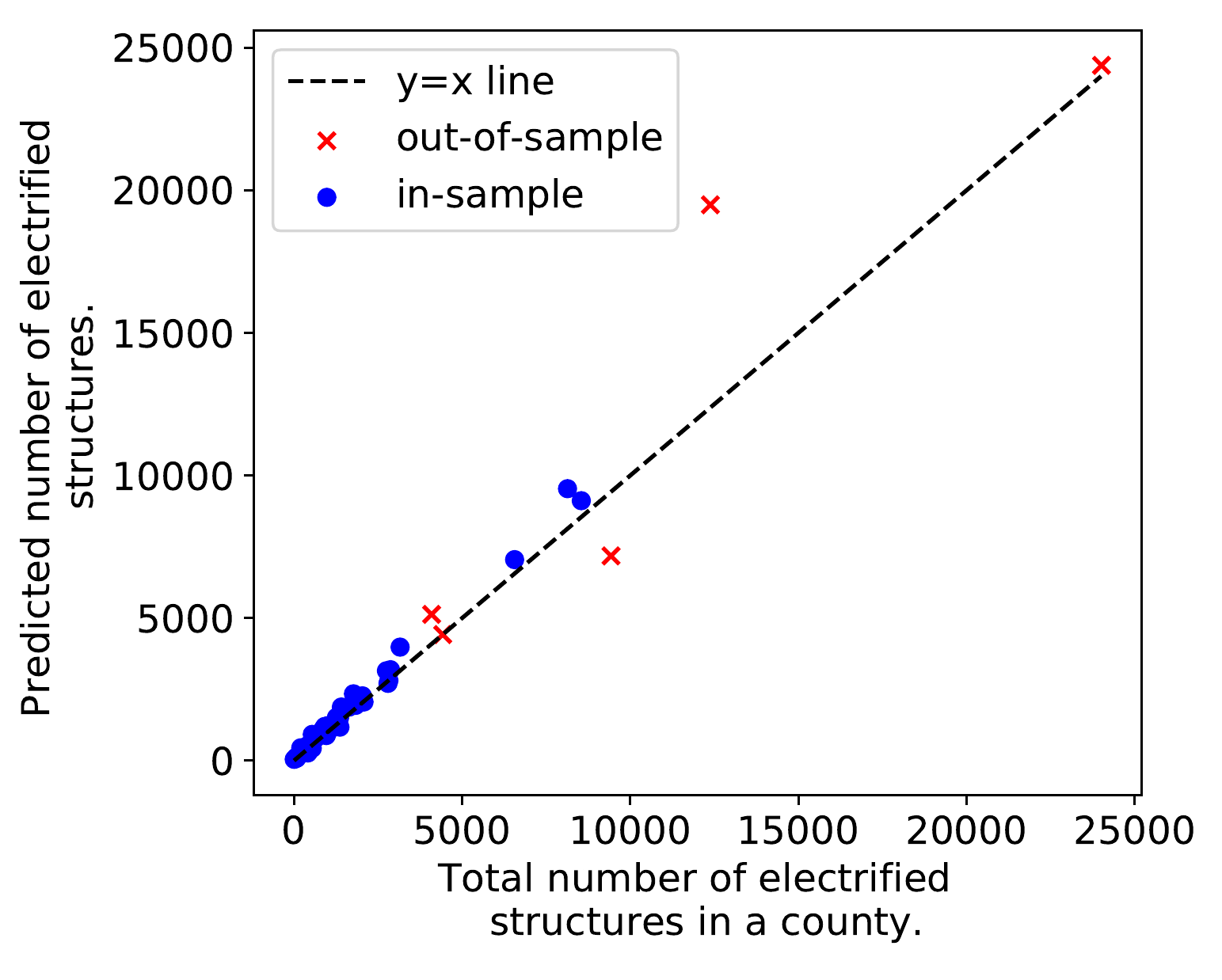} &
\includegraphics[width=0.48\linewidth, bb=0 0 440 440]{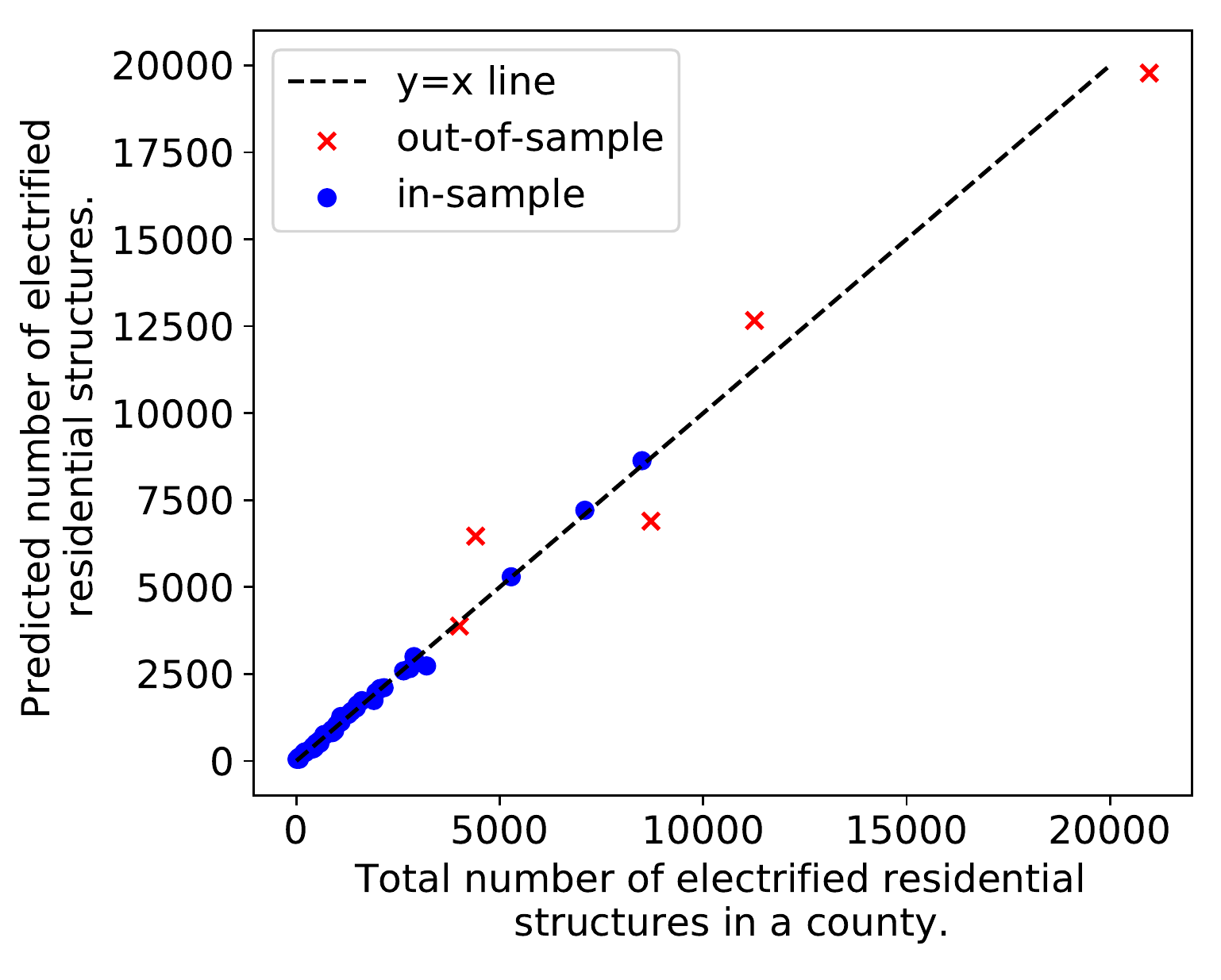} \\
(a) number of electrified structures & (b) number of electrified residential structures\\
($R^2:$ in-sample=0.96, out-of-sample=0.79) & ($R^2:$ in-sample=0.98, out-of-sample=0.94)\\
\end{tabular}
\caption{Regressions for in-sample and out-of-sample test sets. Each dot represents a county. County-level $R^2$ values were computed by comparing county-level predicted and actual total values.}
\label{regression_figure}
\end{figure}

We present the performance results of models trained for the three prediction objectives discussed in Section \ref{sec:objectives}. Note that for classification models, we report performance as overall and category-level accuracy values, and for regression models we report performance using $R^2$ scores. 

\textbf{Objective A. Measuring electricity access:} We first evaluate the performance of our 3-class electrification access model against a baseline -- nighttime lights based High Resolution Electricity Access (HREA) dataset \cite{hrea}, in the out-of-sample counties, as shown in Table \ref{tab:objective1_results}(a). For consistent comparison with HREA, the 3-class output of our model -- no, unelectrified, and electrified structures -- was converted to a binary output by grouping no buildings and unelectrified structures. Our model performs significantly better than the baseline in identifying areas with electricity access. Better detection performance could be attributed to high resolution of daytime satellite images that hold more visual information about structures and surrounding areas, which is lacking in low resolution NL images. Although both models perform comparably in detecting unelectrified areas, our model provides further differentiation between areas with unelectrified structures or no structures at all. Moreover, Table \ref{tab:objective1_results}(b) shows that our 3-class model exhibits strong overall and category-level performance in classifying electricity access categories in both the in-sample and out-of-sample test sets. 

\textbf{Objective B. Quantifying the extent of electrification:} Table \ref{tab:objective2_results} summarizes results for predicting the percent and number of electrified structures. The binary-classification model shows strong overall performance in identifying (low/high) percent of electrified structures in both test sets. However, the model tends to perform relatively better in identifying regions with a low percentage of electrified structures and this behavior is more evident in the out-of-sample test set. Many low electrification regions tend to be rural, which may in turn make detection easier for our model. Our regression model for predicting the number of electrified structures shows fairly strong overall results for the in-sample test ($R^2=0.78$) and even though it degrades slightly on the more challenging out-of-sample test ($R^2=0.66$), these results are indicative of our model's ability to generalize well. Furthermore, for each county in the test sets, we observed a strong agreement between the total predicted number and the actual number of electrified structures (See Figure \ref{regression_figure}(a)). 

\textbf{Objective C. Quantifying types of electrified structures:} Table \ref{tab:objective3_results} summarizes results for predicting the percent and number of residential electrified structures. Our classification model yields reasonably good results on both test sets and shows consistent performance across both classes -- (low/high) proportions of electrified residential structures. Although we observe promising regression results for predicting the number of electrified residential buildings in the in-sample counties ($R^2=0.80$), model performance in out-of-sample counties is modest ($R^2=0.55$). We found that our regression model performed worse in one out-of-sample county (from the top-right quadrant of Kenya as shown in Figure \ref{images_map}(b)), which decreased the overall $R^2$ from $0.65$ (calculated using the remaining 4 counties) to $0.55$. Further investigation showed that the total number of observations from the top-right quadrant of Kenya was lower than the other quadrants resulting in poorer performance in the out-of-sample county from that quadrant. It indicates that our model struggles to generalize and needs further improvement. However, we do observe a strong agreement between the predicted and actual number of electrified residential structures when aggregated at the county-level (Figure \ref{regression_figure}(b)).

\section{Conclusion and Future Work}
\label{sec:conclusion}

Based on our initial results, we propose the following extensions. First, one approach to improve the generalizability of our regression models in unseen areas would be to train our models on a building segmentation task to explicitly establish a relationship between building locations and counts. We also plan to include multiple classes in our classification models and validate our approach in other Sub-Saharan African countries with similar ground truth datasets. Future work will also include in-depth explanations of learnt features given neural activations in our models.

Spatially-consistent, high-resolution data on electrification access and rate of electrification are crucial to more efficient and targeted infrastructure investments towards achieving universal energy access. Data on density of different types of structures can help energy system planners and policy makers identify areas with different energy demand requirements, enabling better estimation of future demand for new electricity systems. In this work we demonstrate approaches to monitor and inform progress toward SDG7. Given our strong prediction model performances, our work can support continuous measurement of electrification at scale in developing countries.

\bibliographystyle{plain}
\bibliography{references}

\begin{thebibliography}{10}

\bibitem{gridfinder}
Christopher Arderne, Conrad Zorn, Claire Nicolas, and EE~Koks.
\newblock Predictive mapping of the global power system using open data.
\newblock {\em Scientific data}, 7(1):1--12, 2020.

\bibitem{usu}
Jacob Arndt and Dalton Lunga.
\newblock Large-scale classification of urban structural units from remote
  sensing imagery.
\newblock {\em IEEE Journal of Selected Topics in Applied Earth Observations
  and Remote Sensing}, 14:2634--2648, 2021.

\bibitem{wb_elecaccess}
The~World Bank.
\newblock Access to electricity (\% of population), 2019.

\bibitem{solarPV}
Kyle Bradbury, Raghav Saboo, Timothy~L Johnson, Jordan~M Malof, Arjun
  Devarajan, Wuming Zhang, Leslie~M Collins, and Richard~G Newell.
\newblock Distributed solar photovoltaic array location and extent dataset for
  remote sensing object identification.
\newblock {\em Scientific data}, 3(1):1--9, 2016.

\bibitem{littlelight}
Santiago Correa, Zeal Shah, and Jay Taneja.
\newblock This little light of mine: Electricity access mapping using
  night-time light data.
\newblock In {\em Proceedings of the Twelfth ACM International Conference on
  Future Energy Systems}, e-Energy '21, page 254–258, New York, NY, USA,
  2021. Association for Computing Machinery.

\bibitem{eog_reliability}
Christopher~D Elvidge, Feng-Chi Hsu, Mikhail Zhizhin, Tilottama Ghosh, Jay
  Taneja, and Morgan Bazilian.
\newblock Indicators of electric power instability from satellite observed
  nighttime lights.
\newblock {\em Remote Sensing}, 12(19):3194, 2020.

\bibitem{gdessa}
Giacomo Falchetta, Shonali Pachauri, Simon Parkinson, and Edward Byers.
\newblock A high-resolution gridded dataset to assess electrification in
  sub-saharan africa.
\newblock {\em Scientific data}, 6(1):1--9, 2019.

\bibitem{DG}
Digital Globe.
\newblock Basemap + vivid product, 2020.

\bibitem{usu_nigeria}
Jordan Graesser, Anil Cheriyadat, Ranga~Raju Vatsavai, Varun Chandola, Jordan
  Long, and Eddie Bright.
\newblock Image based characterization of formal and informal neighborhoods in
  an urban landscape.
\newblock {\em IEEE Journal of Selected Topics in Applied Earth Observations
  and Remote Sensing}, 5(4):1164--1176, 2012.

\bibitem{KNES}
The World~Bank Group.
\newblock Kenya national electrification strategy: Key highlights, 2018.

\bibitem{msftbuild}
Mehdi~P Heris, Nathan~Leon Foks, Kenneth~J Bagstad, Austin Troy, and Zachary~H
  Ancona.
\newblock A rasterized building footprint dataset for the united states.
\newblock {\em Scientific Data}, 7(1):1--10, 2020.

\bibitem{hrsl}
Facebook~Connectivity Lab and Center for International Earth Science
  Information Network CIESIN Columbia~University.
\newblock High resolution settlement layer (hrsl), 2016.

\bibitem{deeproof}
Stephen Lee, Srinivasan Iyengar, Menghong Feng, Prashant Shenoy, and Subhransu
  Maji.
\newblock Deeproof: A data-driven approach for solar potential estimation using
  rooftop imagery.
\newblock In {\em Proceedings of the 25th ACM SIGKDD International Conference
  on Knowledge Discovery \& Data Mining}, pages 2105--2113, 2019.

\bibitem{hrea}
Brian Min and Zachary O'Keeffe.
\newblock High resolution electricity access indicators dataset, 2021.

\bibitem{crowdsource_grid}
Pezhman Nasirifard, Jose Rivera, Qunjie Zhou, Klaus~Bernd Schreiber, and
  Hans-Arno Jacobsen.
\newblock A crowdsourcing approach for the inference of distribution grids.
\newblock In {\em Proceedings of the Ninth International Conference on Future
  Energy Systems}, e-Energy '18, page 187–199, New York, NY, USA, 2018.
  Association for Computing Machinery.

\bibitem{VGG}
Karen Simonyan and Andrew Zisserman.
\newblock Very deep convolutional networks for large-scale image recognition.
\newblock {\em arXiv preprint arXiv:1409.1556}, 2014.

\bibitem{googlebuild}
Wojciech Sirko, Sergii Kashubin, Marvin Ritter, Abigail Annkah, Yasser
  Salah~Eddine Bouchareb, Yann~N. Dauphin, Daniel Keysers, Maxim Neumann,
  Moustapha Ciss{\'{e}}, and John Quinn.
\newblock Continental-scale building detection from high resolution satellite
  imagery.
\newblock {\em CoRR}, abs/2107.12283, 2021.

\bibitem{daytime_energy_consumption}
Artem Streltsov, Jordan~M Malof, Bohao Huang, and Kyle Bradbury.
\newblock Estimating residential building energy consumption using overhead
  imagery.
\newblock {\em Applied Energy}, 280:116018, 2020.

\bibitem{india_microgrids}
Kush~R Varshney, George~H Chen, Brian Abelson, Kendall Nowocin, Vivek Sakhrani,
  Ling Xu, and Brian~L Spatocco.
\newblock Targeting villages for rural development using satellite image
  analysis.
\newblock {\em Big Data}, 3(1):41--53, 2015.

\end{thebibliography}

\end{document}